\documentclass[runningheads]{llncs}
\usepackage[T1]{fontenc}
\usepackage{graphicx}
\usepackage{hyperref}
\usepackage{cite}
\usepackage{tabularx}
\usepackage{eurosym}
\usepackage{float}

\usepackage[table]{xcolor}
\newcommand{\hlPIa}[1]{\colorbox{red!20}{#1}}     % PI032
\newcommand{\hlPIb}[1]{\colorbox{red!20}{#1}}  % PI027
\newcommand{\hlTGg}[1]{\colorbox{green!20}{#1}}   % TG23
\newcommand{\hlTGb}[1]{\colorbox{green!20}{#1}}    % TG16
\newcommand{\hlTGp}[1]{\colorbox{green!20}{#1}}  % TG29
\newcommand{\hlTHy}[1]{\colorbox{blue!15}{#1}}  % TH89

\begin{document}
\title{From Policy Documents to Structured Survey Responses: Evaluating Large Language Models for Policy Monitoring}
\titlerunning{AI-Driven Survey Responses in STIP Compass}
%\author{Carolyn Cole\inst{1}\orcidID{0000-1111-2222-3333} \and
%Second Author\inst{2,3}\orcidID{1111-2222-3333-4444} \and
%Third Author\inst{3}\orcidID{2222--3333-4444-5555}}
\author{Carolyn Cole \and
Matthias Deschryvere \and
Toqeer Ehsan \and
Arash Hajikhani}
\authorrunning{Cole et al.}
\institute{Reliable Intelligence Team, VTT Technical Research Centre of Finland Ltd., 02150 Espoo, Finland\\
\email{\{firstname.lastname\}@vtt.fi}\\
}
\maketitle              % typeset the header of the contribution
\begin{abstract}
Science, technology, and innovation policies are crucial for competitiveness, yet their diversity and scale make them difficult to map and monitor consistently. Existing approaches rely heavily on manual survey efforts, which are costly and challenging to scale across countries. Large language models (LLMs) enable new possibilities for extracting and structuring information from long and unstructured policy documents.
This paper presents an application of LLMs as “AI respondents” for generating structured survey responses from policy texts. We develop a data extraction pipeline based on long-context in-context learning to map information from public web sources into predefined survey categories, including policy instruments, target groups, and thematic areas. The pipeline integrates a validation step using a secondary LLM to assess relevance and evidence, alongside comparisons with human-provided responses.
Using a multi-country dataset, we evaluate the alignment between LLM-generated and human-generated outputs through overlap measures and cross-validation. Results show that LLMs achieve high agreement for structured indicators (84–95\%), while differences remain in free-text fields, where models tend to provide more detailed procedural descriptions.
These findings highlight the potential of hybrid human–AI workflows for policy monitoring, improving both efficiency and scalability while maintaining the need for human validation and contextual interpretation.

\keywords{LLMs \and Policy Intelligence \and AI Respondents \and Long-context In-context Learning \and Survey Automation \and Information Extraction.}
\end{abstract}

\section{Introduction}
\label{sec:1}

Science, Technology, and Innovation (STI) policies are complex socio-technical constructs that play a central role in shaping national competitiveness and addressing global challenges. Yet, systematic mapping and continuous monitoring of these policies remain costly and labor-intensive, particularly in the context of large-scale international surveys such as the EC-OECD STIP Compass \cite{Flanagan2011}. The Compass aggregates data on STI policy initiatives across OECD and partner countries, relying on expert respondents to fill in structured survey instruments linked to web-based sources of evidence. While this approach provides a unique comparative perspective, it faces challenges of scale, consistency, and timeliness as the number and complexity of initiatives expand.

LLMs are redefining natural language processing (NLP) by enabling machines to internalize knowledge from large unstructured corpora and to adapt to diverse downstream tasks through prompting rather than parameter updates \cite{tan2402large, mao2025survey, dherin2025learning}. Recent generative LLMs such as GPT-4o are capable of long-context reasoning and in-context learning, making them suitable for information extraction from extended policy documents and web content. Their ability to generate structured responses aligned with human-designed schemas offers a potential solution to the persistent difficulties of innovation policy data collection and validation.

However, integrating LLMs into international policy monitoring is not straightforward. Prior work shows that LLMs can act as ``artificial respondents'', replicating social science experiments by generating survey answers conditioned on demographic profiles \cite{Ashokkumar2024}. On the other hand, LLM-driven data generation may introduce systematic biases and distortions—so-called model collapse—if models are repeatedly trained on synthetic outputs \cite{Shumailov2024}. Moreover, innovation policy data pose unique challenges as policies are heterogeneous, multi-scalar, and embedded in institutional contexts that are not always captured in publicly available sources \cite{Feldman2015}.

In the STIP Compass workflow, expert respondents from member nations cite web-based responses for each policy initiative and fill a structured survey by selecting taxonomy codes \cite{ec-oecd-2025-stip-taxonomy} for policy instruments (PI), target groups (TG), and policy themes (TH) along with free-text fields. A short example of this workflow is shown in Table~\ref{tab:0}. We cast this as an LLM-assisted survey-filling task: given the web-scraped initiative text, the model predicts the corresponding sets of PI/TG/TH codes and generates the free-text fields: description and objectives.

\begin{table}[ht]
\small
\centering
\caption{\small A sample workflow for web-based STIP Compass survey filling.}
\label{tab:0}
\begin{tabularx}{\linewidth}{lX}
\hline
%Web text & ...The \textbf{Government} is increasingly looking to utilize \textbf{artificial intelligence} to make, or assist in making, \textbf{administrative decisions}… compatible with core administrative law principles such as \textbf{transparency, accountability, legality, and procedural fairness}.
%The objective of this Directive is to ensure that \textbf{Automated Decision Systems are deployed} in a manner that \textbf{reduces risks to Canadians and federal institutions}…
%…\textbf{Completing an Algorithmic Impact Assessment} prior to the production of any Automated Decision System… \textbf{Applying the relevant requirements} prescribed in…
%…developing processes so that the data and information used … are \textbf{tested for unintended data biases}…
%…\textbf{providing a meaningful explanation to affected individuals} of how and why the decision was made…
%…\textbf{Contracted third-party vendor} with a related specialization…
%…Data and information on the use of Automated Decision Systems … are made available to the \textbf{public}, where appropriate... \\

Web text & The \hlTGg{Government} is increasingly looking to utilize
\hlTHy{artificial intelligence} to make, or assist in making, administrative decisions\dots
compatible with core administrative law principles such as \hlTHy{transparency, accountability, legality, and procedural fairness.}
The objective of this \hlPIa{Directive} is to ensure that Automated Decision Systems are deployed in a manner that reduces risks to \hlTGb{Canadians} and \hlTGg{federal institutions} \dots \hlPIb{Completing an Algorithmic Impact Assessment} prior to the production of any Automated Decision System \dots \hlPIa{Applying the relevant requirements prescribed in} \dots developing processes so that the data and information used \dots are
\hlPIb{tested for unintended data biases} \dots \hlPIa{providing a meaningful explanation} to \hlTGb{affected individuals} of how and why
the decision was made \dots \hlTGp{Contracted third-party vendor} with a related specialization \dots Data and information on the use of Automated Decision Systems\dots are made available to the \hlTGb{public,} where appropriate...\\
\hline
Codes & PIs: PI027, PI032, TGs: TG16, TG23, TG29, THs: TH89\\
\hline
Text fields & Description: Federal policy instrument providing a risk-based approach...\\
 &  Objectives: Automated decision systems deployed by federal...\\
\hline
Label defs. & \hlPIa{PI027}: Governance | Standards and certification for technology development and adoption \\
& \hlPIa{PI032}: Guidance, regulation and incentives | Science and technology regulation and soft law\\
& \hlTGb{TG16}: Social groups especially emphasised | Civil society\\
& \hlTGb{TG23}: Governmental entities | National government\\
& \hlTGb{TG29}: Firms by size | Firms of any size\\
& \hlTHy{TH89}: Research and innovation for society | Ethics of emerging technologies\\
\hline
\end{tabularx}
\end{table}

This paper contributes to the emerging field of AI-assisted policy intelligence by presenting an operational pipeline that integrates LLM capabilities for the EC-OECD STIP Compass.
Specifically, we design and test a data extraction pipeline that uses long-context prompting to map survey taxonomy codes (policy instruments, themes, and target groups) from web-scraped content provided by survey respondents. A secondary validation layer employs an LLM to evaluate outputs on dimensions of relevance and evidence. Using a pilot across six OECD countries (Canada, Finland, Germany, Korea, Spain, and Türkiye), we assess the overlap between LLM-generated and human-generated survey responses and explore complementarities in descriptive and objective fields \cite{Hajikhani2024VTT}. Our findings show that LLMs achieve high overlap in structured codes (84–95\%) but diverge in textual fields, where AI tends to provide more detailed procedural descriptions while humans emphasize contextual and societal impacts. These insights highlight the promise of hybrid human-AI approaches for international policy monitoring. The key contributions of this paper are as follows:
\begin{itemize}
    \item We design an integrated data extraction pipeline that leverages long-context in-context learning to process lengthy unstructured policy documents.
    \item We implement a validation layer that evaluates relevance and evidence by employing another LLM as a validator model.
    \item We evaluate the pipeline in a pilot study covering six OECD countries, analyzing overlap, agreement, and cross-validation between LLM-generated and human-generated responses.
\end{itemize}

\section{Related Work}
\label{sec:2}

The methodological challenges of collecting and comparing STI policy data have long been recognized in the literature on policy mixes and innovation systems. Policies are difficult units of analysis, and large-scale cross-country data are costly to compile and validate \cite{Flanagan2011}. Efforts to address these challenges have included international surveys and expert-driven databases such as the OECD STIP Compass, yet these approaches are constrained by reporting burden, data gaps, and inconsistencies across national contexts.

The rise of LLMs introduces new opportunities to address these challenges. LLMs have been applied successfully in tasks such as information extraction, summarization, and question answering, often outperforming earlier supervised NLP methods \cite{tan2402large, mao2025survey}. Their in-context learning capabilities allow them to adapt dynamically to survey-style questions without the need for costly labeled training data \cite{dherin2025learning}. Recent studies demonstrate the ability of LLMs to act as proxies for human subjects in social science experiments, suggesting their potential as scalable substitutes or complements to traditional survey respondents \cite{Ashokkumar2024}.

At the same time, concerns remain about their reliability. Shumailov et al. \cite{Shumailov2024} warn of distributional drift and degradation in model outputs when systems recursively train on synthetic data. In the context of STI policy, the absence of gold-standard labeled datasets and the heterogeneous nature of policy initiatives make fine-tuning approaches less feasible, as highlighted in recent experimentation with the STIP Compass \cite{Hajikhani2024VTT}. Instead, long-context prompting combined with expert-designed taxonomies offers a pragmatic way to leverage LLMs while maintaining human oversight.

Our work builds on these strands by testing an operational pipeline that integrates LLMs into the STIP Compass survey process. While prior research has explored web-based policy document analysis and retrieval-augmented methods, our contribution is to compare human-provided and AI-generated responses across multiple dimensions of STI policy data. In doing so, we extend calls to leverage the ``new data frontier'' in innovation studies \cite{Feldman2015} through AI-driven approaches for international policy monitoring.
%Our work builds directly on these strands by testing an operational pipeline that integrates LLMs into the STIP Compass survey process. While prior research has explored web-based policy document analysis and retrieval-augmented methods, the contribution here is to demonstrate, for the first time, a systematic comparison between human-provided survey responses and AI-generated responses across multiple dimensions of STI policy data. In doing so, we extend earlier calls to leverage the ``new data frontier" in innovation studies \cite{Feldman2015} with AI-driven approaches that are both scalable and adaptable to international policy monitoring.

\section{Methodology}
\label{sec:3}
\begin{figure}[h]
\centering
\includegraphics[width=1.0\textwidth]{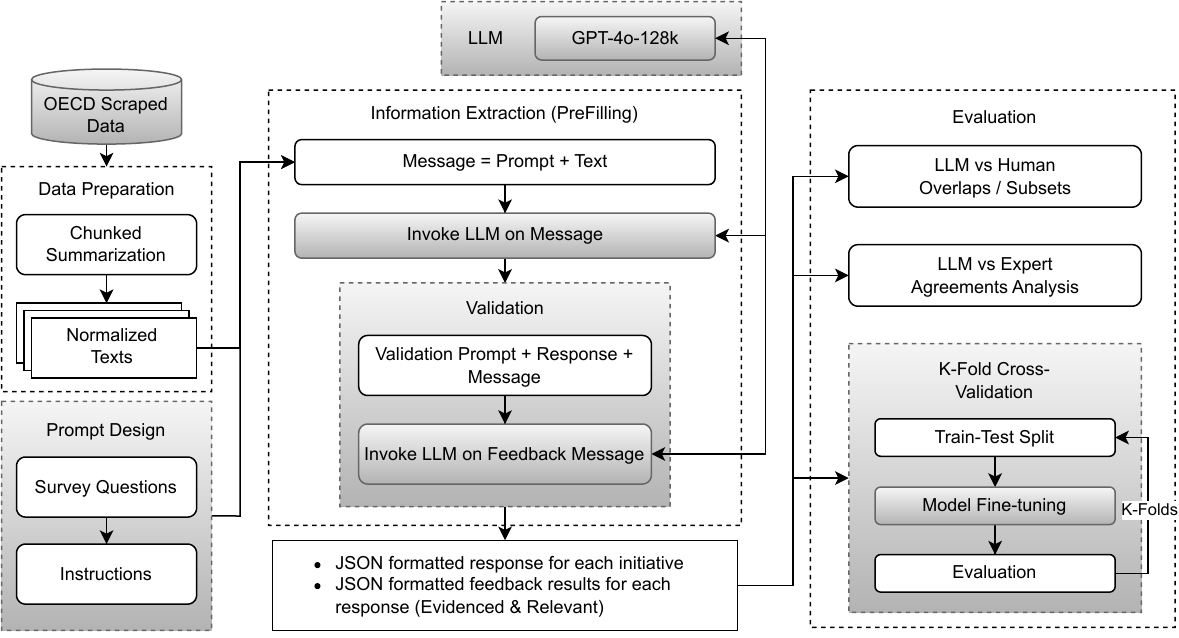}
\caption{\small Methodological workflow of the study: starting from OECD-scraped policy texts, followed by chunked summarization and data preparation, prompt design with survey questions, and information extraction using GPT-4o-128k \cite{openai2024gpt4o}. Validation and evaluation involve comparing LLM-generated outputs with human annotations, as well as cross-validation using model fine-tuning.}
\label{fig:1}
\end{figure}

In this section, we describe the data extraction pipeline for the EC-OECD STIP Compass survey. The raw data were obtained from the OECD and consist of the content from URLs that survey participants identified as relevant policy initiatives. The following subsections describe the preparation of the data for pre-filling, prompt design, and evaluation. Figure~\ref{fig:1} illustrates the workflow of our methodology.

\subsection{Data Preparation}
\label{sec:3-1}
We processed the survey data and the scraped text provided by the OECD to retain only those initiatives with sufficient content for analysis. Initiatives containing fewer than 200 tokens (fewer than 100 words) were discarded as insufficient for analysis. In contrast, initiatives with more than 120,000 tokens were further processed to fit within the LLM context window. For this purpose, we devised a chunked summarization method that divided the initiative content into chunks of 50,000 tokens and prompted an LLM to summarize each chunk while retaining underlying information related to STI policies. The resulting summaries were then aggregated by the LLM to produce a complete text containing all relevant initiative information. The prompts designed for the chunked summarization method are given below.

\begin{quote}
%\small
\textit{Individual Chunk Prompt: ``The following is a set of texts related to a policy initiative: ``+ chunk +'' Based on this list of docs, write a detailed account covering description, objectives, dates, policy themes and instruments, key stakeholders, budgets, and evaluations. Capture exact details, examples, and cases for these dimensions. Be thorough, detailed, and comprehensive. Use only text from the provided documents.''}
\end{quote}

\begin{quote}
%\small
\textit{Chunk Summary Aggregation Prompt: ``The following is a set of detailed summaries: ``+ chunked\_summaries +'' Take these and generate a detailed account covering description, objectives, dates, policy themes and instruments, key stakeholders, budgets, and evaluations. Capture exact details, examples, and cases for these dimensions. Be thorough, detailed, and comprehensive. Use only text from the provided summaries.''}
\end{quote}

\subsection{Pre-filling - Long-Context In-Context Learning}
\label{sec:3-2}
Long-context in-context learning refers to the ability of LLMs with extended context windows to generalize examples embedded in lengthy prompts, often spanning thousands of tokens \cite{bertsch-etal-2025-context}. Retrieval-augmented generation (RAG), in contrast, integrates an LLM with an external retrieval system, typically querying a document index with a neural retriever and prompting the model to generate output based on retrieved content \cite{Lewis2020rag, asai2023selfrag}. 
We did not adopt RAG for survey pre-filling from long STI policy documents for two reasons: first, the knowledge to be extracted from scraped policy content was not explicitly defined, making it difficult to formulate direct fact-based queries; second, adapting RAG by splitting survey questions into smaller prompts leads to redundancy, as overlapping policy indicators and guidelines produce overlapping content.
%Through selective manual qualitative analysis, we found that RAG performs suboptimal for survey Pre-filling from long STI policy documents because; 1) the knowledge to be extracted from scraped policy content was not explicitly defined, making it difficult to formulate direct fact-based queries; and 2) adapting RAG by splitting survey questions into smaller prompts leads to redundancy, as overlapping policy indicators and guidelines produce overlapping content. %; (3) the fragmentation of survey questions significantly increases the number of API calls requiring high resource utilization.

Long-context in-context learning enables the inclusion of complete content, survey questions, and extraction guidelines in a single extended prompt. We adopted this approach to preserve coherence, incorporate examples, and capture relevant material without restricting nuanced findings. The context window accommodated most cases in our dataset.
%Long-context ICL enables the inclusion of complete content, survey questions, and extraction guidelines in a single extended prompt, leveraging the long context windows in the latest LLMs to preserve coherence and yield detailed responses. We adopted a long-context ICL that incorporated examples within the prompts to shape and improve the quality of responses. This approach captures a wide yet relevant range of material without restricting nuanced findings, making it well-suited for our study. The capacity of current large context window LLMs adequately accommodated most of the cases in our dataset. 
%Only six initiatives had content having more than 120,000 tokens and were preprocessed using fragmented summarization. 
%The long-context ICL allows for more nuanced and adaptable responses, which are crucial for the extraction of information from long unstructured STI policy initiatives.

\subsubsection{Prompt Design}
We incorporated survey questions and OECD guidelines into the prompt design. The prompts covered descriptions, objectives, policy instruments, target groups, policy themes, start date, budget, and evaluation report. They instructed the model to respond in English, rely only on the provided content, and return structured outputs for easier parsing and integration. The full prompts are provided in Appendix~\ref{app:a-3}.
%We incorporated survey questions and OECD guidelines into the prompt design. %The iterative approach involves testing each prompt and validating the resulting responses for quality, followed by further adjustments and improvements. 
%The primary objective of the prompt design was to ensure responses in English while accommodating multilingual cases. The guidelines covered all indicators, including descriptions and objectives, policy instruments (PI), target groups (TG) policy themes (TH), start date, budget, and evaluation report. We experimented with various arrangements of instructions, classifications, and examples in the prompts to ensure consistent and unified responses. Additionally, we designed the prompts to produce responses in a structured format for easier parsing and integration.

%We adhered to a schema in which the LLM provided the context and predefined classifications, particularly for policy instruments, target groups, and policy themes. We ensured that the information identified from the raw scraped text was appropriately categorized. The prompt instructions emphasized avoiding the generation of new content and instead relying on the provided content. The list of designed prompts is provided in Appendix~\ref{app:a-3}. %Additional safeguards are implemented to maintain coherency and avoid verbose and excessive jargon, ensuring consistency in responses.

\subsubsection{Response Validation}
LLM-based evaluation has become common for assessing generated outputs on dimensions such as groundedness, completeness, and relevance \cite{gu2024survey, li2024llms, saha2025learning, kim2023better}.
%Evaluation metrics have evolved with the growing popularity of LLMs \cite{gu2024survey, li2024llms}. A notable development is the use of LLM-based evaluation, where one model employs Chain-of-Thought (CoT) reasoning to assess the outputs of another LLM \cite{saha2025learning}. This approach emphasizes dimensions such as groundedness and completeness through well-designed prompting strategies, which can then be scored numerically \cite{kim2023better}. 
Research indicates that larger model sizes generally produce improved performance in summarization evaluation, with stronger correlation to human judgments \cite{liu2023g, fabbri2021summeval}. In addition, evaluation methods may employ reference-based approaches that compare the generated text with the ground truth or reference texts \cite{wu2023large}. %For tasks without a single correct answer, LLM-based evaluation metrics such as GPT-4-based G-Eval have been shown to align better with human quality assessments than traditional reference metrics \cite{liu2023g}. However, constructing high-quality human reference datasets remains costly and resource intensive.

To validate the generated responses, we employed another instance of the LLM to evaluate them against the prompt and source material. A binary scoring scheme assessed \textit{evidence} and \textit{relevance}, indicating whether each response was supported by the text and followed the instructions. The structured validation prompt is provided in Appendix~\ref{app:a-1}. This step filtered out cases that could have resulted from hallucinations or misinterpretations.
%To validate the generated responses, we employed another instance of the LLM to evaluate them against the prompt and source material. A binary (0/1) scoring scheme was applied to assess two factors: \textit{evidence} and \textit{relevance}. The model evaluates whether each response is evidenced by the text and whether the response is relevant to the instructions given. We developed a structured evaluation prompt (\ref{app:a-1}) to ensure a consistent and objective assessment of the generated responses, focusing on their adherence to the source material and relevance to the instructions. This evaluation filtered out cases that could have resulted from hallucinations or misinterpretations. 
%To further reduce risks of bias and hallucination, the pipeline constrains the LLM to predefined STIP Compass taxonomies and instructs it to rely only on the provided source text. The validation step also checks whether generated responses are evidenced and relevant, while human comparison is used to identify cases where model outputs diverge from expert responses.

\subsection{Evaluations}
In addition to response validation, we incorporated three further evaluation measures. 
\begin{enumerate}
    \item \textbf{Overlap analysis} compared human- and LLM-generated survey responses, capturing the extent of alignment between the two datasets.

    \item \textbf{Label-wise agreement scoring} quantified consistency between human annotations and model outputs for policy labels using high-, medium-, and low-agreement categories.

    \item \textbf{K-fold cross-validation} validated LLM-generated labels using fine-tuned masked and causal models, addressing the absence of a full gold-standard reference for all structured policy indicators.
\end{enumerate}

%\begin{enumerate}
%\item \textit{Overlap analysis} provided a comparison between human- and LLM-generated survey responses, capturing the extent of alignment between the two datasets.
%\item The naïve overlap calculation does not necessarily provide a meaningful insight into agreement, as it only reflects surface-level similarity. Therefore, \textit{label-wise agreement} scoring is applied to quantify the consistency between the human annotations and model outputs with respect to policy labels using high-, medium-, and low-agreement categories.
%\item Manual extraction of the structured policy indicators from long, unstructured documents is labor-intensive and prone to error. Moreover, the LLM-generated survey responses could not be fully validated due to the absence of gold-standard reference. To address this limitation, we validated LLM-responses against structured policy labels using \textit{k-fold cross-validation} by fine-tuning masked and causal models.
%\end{enumerate}

\section{Experimental Setup}
\label{sec:4}
In our experiments, we performed survey pre-filling using LLMs as survey respondents to extract and generate multiple types of survey fields. Our study addressed two key questions: (1) Can web-scraped content provide sufficient and relevant information to pre-fill STIP Compass survey questions? (2) Is it feasible to map unstructured web-scraped content to structured survey categories using LLMs to generate survey responses in place of human respondents? 
We hypothesized that long-context LLMs could capture a significant portion of structured information for free-text fields as well as multi-label policy identification. 

\subsection{Choice of LLM}
We selected GPT-4o-128k \cite{openai2024gpt4o} for our study because of its extended context window and strong performance across evaluation benchmarks. The selection was supported by the latest metrics from Stanford's Holistic Evaluation of Language Models (HELM), which provides a comprehensive assessment of language models' capabilities and limitations. %HELM evaluates language models on multiple dimensions, including accuracy, calibration, robustness, fairness, bias, toxicity, and efficiency, in various scenarios. 
According to the HELM leaderboard\footnote{https://crfm.stanford.edu/helm/lite/latest}, GPT-4o (2024-05-13) achieved a mean win rate of 0.938 on standard evaluation metrics.

%\subsection{Operational LLM Pipeline}
%Our operational methodological pipeline employs GPT-4o 128k with low variability in responses and relevant evidence for performance. The process includes a validation layer that assesses LLM performance (e.g., flagging hallucinations). Finally, data from survey responses are obtained and incorporated to create a comparison with survey participant responses, allowing validation during the analysis iteration. Figure~\ref{fig:2} illustrates our LLM-based methodological pipeline.

%\subsubsection{Validation}
%To validate the generated responses, we employed another instance of GPT-4o-128k to evaluate the responses against the prompt and raw material. Using a structured evaluation prompt (\ref{app:a-1}), our objective was to ensure a consistent and objective evaluation of the generated responses, focusing on their adherence to the source material and relevance to the instructions given. This evaluation filters out the cases where the LLM response could be the result of a hallucination or misinterpretation.

\subsection{Evaluation Design}
To validate the generated responses, we employed another instance of GPT-4o-128k to evaluate the responses against the prompt and raw material. We used a structured evaluation prompt (\ref{app:a-1}) to ensure consistency in assessing the generated responses, focusing on their adherence to the source material and relevance to the instructions. 
%This evaluation filtered out the cases where the LLM response could have resulted of a hallucination or misinterpretation.

For post-extraction evaluations, we again employed GPT-4o-128k as an evaluator for free-text fields, including descriptions and objectives. We designed a prompt (\ref{app:a-2}) to compare overlaps and discrepancies between human participants and LLM-generated responses. The prompt instructs the LLM to quantify the results into four categories: full overlap, high overlap, low overlap, and no overlap. However, the degree of overlap against policy labels was quantified using overlap percentages. We computed agreement scores using micro F1 scores throughout the dataset. Similarly, micro F1 scores were used to evaluate k-fold (k=5) cross-validation experiments by fine-tuning (system prompt~\ref{app:a-4}) a range of masked and causal models (Table~\ref{tab:4}).

\subsection{Implementation Details}
Data preparation, pre-filling, response validation, and free-text evaluation were conducted using Azure AI Services\footnote{https://ai.azure.com} with different GPT-4o deployments. 
Dataset analysis and agreement scores were computed using standard Python libraries. In k-fold cross-validation, the dataset was shuffled and split into 80\% training, 10\% validation, and 10\% testing for each fold. 
The main hyperparameters for masked and causal models, as well as LoRA configurations, are summarized in Appendix~\ref{app:a}.

\subsection{Cost}
Running the experiments with GPT-4o incurred a total cost of \EUR{446.84}. This cost applies to the six-country pilot, not full OECD deployment. Full-scale costs would increase with document volume and length, but can be reduced through URL filtering, cached scraping, selective summarization, and targeted human review.

\section{Results \& Discussion}
\label{sec:5}
The results compare human- and LLM-generated survey responses across six OECD countries, focusing on free-text fields and policy indicator labels. The following subsections present human-LLM overlap, label agreement, and k-fold cross-validation results.
%The results of the study provided a systematic comparison between human-generative and LLM-generated survey responses in six sample OECD countries, focusing on free-text fields and policy related indicator labels. The following subsections present Human-LLM overlaps and subsets, agreement analysis of indicator labels, and k-fold cross-validation of LLM-generated labels.

\subsection{Dataset Analysis}
\label{sec:3-3}
%The final dataset contains original unstructured web scraped content with more than 200 tokens and less than 120,000 tokens. All other initiatives have aggregated content from chunked summaries. 
Table~\ref{tab:1} presents the country-level statistics after data preparation and filtering. The \textit{Insufficient} column refers to the share of policy initiatives without URLs or containing less than 200 tokens. The \textit{Unidentified} column presents initiatives that have sufficient content, but relevant policy instruments could not be extracted. The \textit{Sufficient} column shows percentages of initiatives that have sufficient and suitable web content. The last column reports the number of samples with appropriate STI content from policy initiatives. 

\begin{table}[h]
\small
\centering
\caption{\small Distribution of web content and number of samples by country. }
\label{tab:1}
\begin{tabular}{llcccc}
\hline
\textbf{Sr\#}&\textbf{Country} & \textbf{Insufficient} & 
\textbf{Unidentified} & 
\textbf{Sufficient} & 
\textbf{\# of Samples} \\
\hline
1&Canada   & 30\% & 6\%  & 64\% & 149 \\
2&Finland  & 32\% & 11\% & 57\% & 80 \\
3&Germany  & 25\% & 7\%  & 68\% & 193 \\
4&Korea    & 27\% & 20\% & 53\% & 146 \\
5&Spain    & 31\% & 13\% & 56\% & 142 \\
6&Türkiye  & 47\% & 12\% & 41\% & 135 \\
&Total  & -- & -- & -- & 845 \\
\hline
\end{tabular}
\end{table}

Each sample contains eight policy indicators. Policy instruments, target groups, and policy themes include additional sub-labels that refer to underlying STI policies. Table~\ref{tab:2} compares human-generated and LLM-generated label coverage.
%Each sample contains eight policy indicators covering various types of content.
%\textit{Description} and \textit{Objectives} are fields of free text that provide introduction and objectives of STI initiatives. \textit{Start date} indicates when the initiative was launched. 
%\textit{Policy instruments (PI)}, \textit{Target groups (TG)}, and \textit{Policy themes (TH)} contain underlying indicators in the form of sublabels that refer to documented policies. 
%The remaining fields are \textit{Budget} of the initiative and \textit{Evaluation report} if there are any. 
%Table~\ref{tab:2} presents statistics of indicator coverage comparing human-generated and LLM-generated labels. 

\begin{table}[h]
\small
\centering
\setlength\tabcolsep{5pt}
\caption{\small Comparison of label statistics between expert-generated and LLM-generated labels.}
\begin{tabular}{lcccccc}
\hline
\textbf{Generated} & \textbf{PI} & \textbf{Unique PI} & \textbf{TG} & \textbf{Unique TG} & \textbf{TH} & \textbf{Unique TH} \\
\hline
By humans & 1,281 & 27 & 3,828 & 33 & 1,895 & 51 \\
By LLM & 2,336 & 28 & 4,727 & 33 & 3,013 & 57 \\
\hline
\end{tabular}
\label{tab:2}
\end{table}

\begin{figure}[H]
\centering
\includegraphics[width=1.0\textwidth]{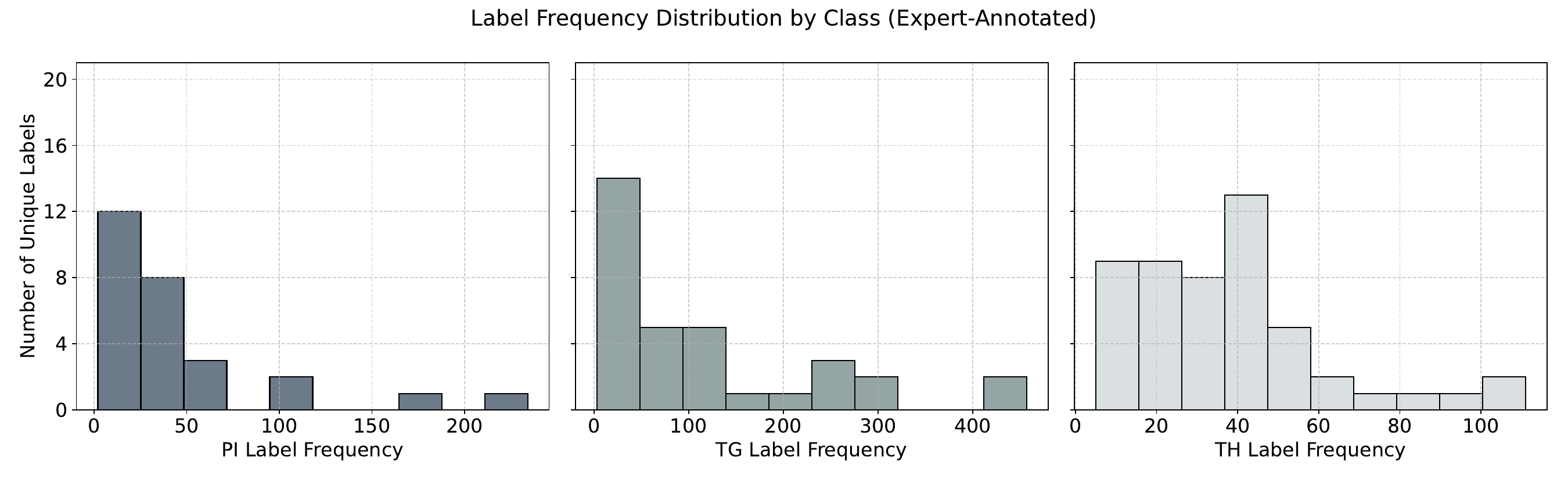}\\[0ex]
\includegraphics[width=1.0\textwidth]{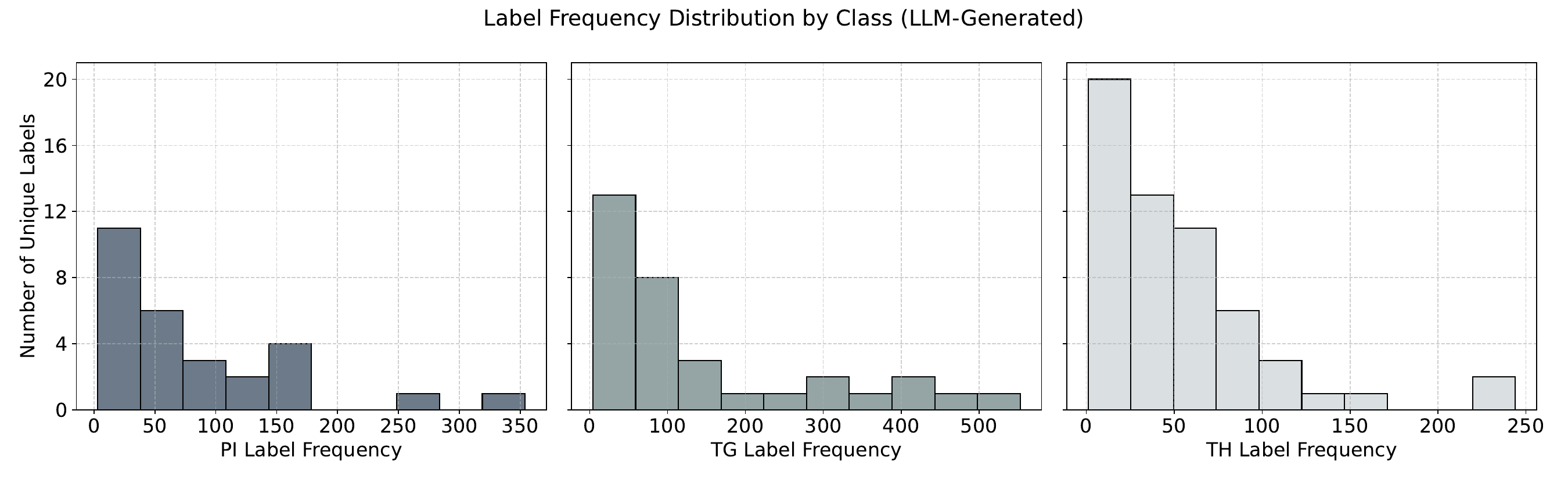}
\caption{\small Comparison of policy indicator labels between expert-generated and LLM-generated labels.}
\label{fig:2}
\end{figure}

Figure~\ref{fig:2} shows the frequency distribution of policy labels by class. Both datasets exhibited similar label distributions, with a significant imbalance across classes. Many labels are underrepresented, with low frequencies in the dataset. %making it challenging to perform cross-validation as an evaluation approach.

\subsection{Human-LLM Overlap}
\label{sec:5-1}
\begin{figure}[ht]
\centering
\includegraphics[width=0.49\textwidth]{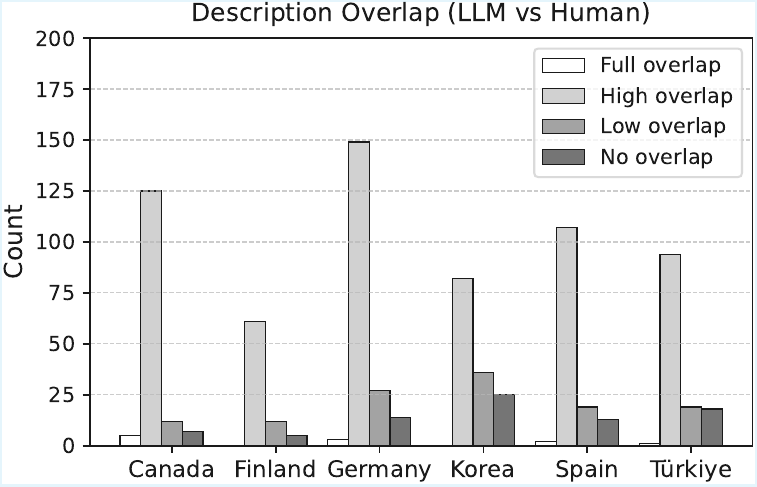}
\includegraphics[width=0.49\textwidth]{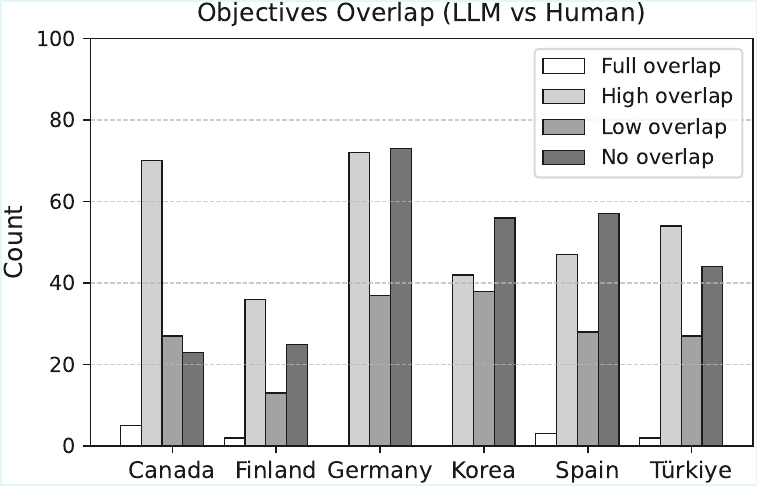}
\caption{\small Overlap of survey participant responses and LLM responses per country on the policy initiative description and objectives.}
\label{fig:3}
\end{figure}

We investigated qualitative differences in free-text responses, identifying complementary tendencies in which LLMs delivered more detailed procedural accounts, while human respondents emphasized contextual and societal dimensions.   
%Similarly to the response validation process, we again employed LLM as an evaluator for free-text fields, description and objectives. We designed a prompt (\ref{app:a-2}) to compare overlaps and discrepancies between human participants and LLM-generated responses. The prompt instructs the LLM to quantify the results into four categories: full overlap, high overlap, low overlap, and no overlap. 
Figure~\ref{fig:3} shows the overlap results for both free-text fields. %The selected text reports the results of comparing the descriptions provided by the participants and the LLMs based on 840 observations. 

The analysis of overlap between descriptions indicated that the predominant share of cases (74.05\%) exhibited high overlap, whereas only 1.19\% demonstrated full overlap, 15.24\% were classified as low overlap, and 9.52\% showed no overlap. These differences suggest that human and AI assessments can complement each other by offering diverse perspectives and insights on the same topics. However, the objective fields showed high overlap in 41\% of the cases, while no overlap was observed in about 36\% of the cases. Low overlap and full overlap were less frequent, occurring in 22\% and 1\% of cases, respectively. These patterns suggest that differences often stem from variations in approach, level of detail, scope, and available information for assessments. 

The divergence in free-text fields appears to reflect differences in emphasis rather than only disagreement. Human responses are generally more concise and contextual, while LLM responses often provide more procedural detail from the scraped text. This is especially visible for objectives, where the LLM may describe implementation steps, whereas human respondents express broader policy aims.

\begin{table}[ht]
\small
\centering
\caption{\small Overlap of survey participant responses and LLM responses per country for policy instruments (A), target groups (B), and policy themes (C).}
\begin{tabular}{llccc}
\hline
& & 
\textbf{Policy instruments} & 
\textbf{Target groups} & 
\textbf{Policy themes} \\
Sr\#&
\textbf{Country}  & 
\textbf{(A)} & 
\textbf{(B)} & 
\textbf{(C)} \\
%\multicolumn{4}{c}{\textit{Share of policy initiatives with overlap between survey answers and LLM answers}} \\
\hline
1& Canada   & 84\% & 97\% & 85\% \\
2& Finland  & 84\% & 98\% & 84\% \\
3& Germany  & 80\% & 97\% & 83\% \\
4& Korea    & 88\% & 93\% & 84\% \\
5& Spain    & 85\% & 94\% & 82\% \\
6& Türkiye  & 88\% & 93\% & 87\% \\
\hline
&Total & 85\% & 95\% & 84\% \\
\hline
\end{tabular}
\label{tab:3}
\end{table}

We further examined overlap in multi-label indicators---policy instruments, target groups, and policy themes---to evaluate LLM outputs relative to human experts.
Table~\ref{tab:3} shows the overlap between labels provided by human participants and those generated by the LLM. The table includes all overlapping cases where at least one policy label overlapped. The LLM performed relatively well in capturing survey respondent codes for policy themes and instruments, and particularly well for policy target groups. On average, in 95\% of policy initiatives the LLM identified at least one of the target groups provided by survey respondents. The corresponding averages are 84\% for policy themes and 85\% for policy instruments. Although some variation exists across countries, these differences are generally limited. %Section~\ref{sec:5-2} provides the label-wise analysis agreement between the human response and the LLM-generated labels.

\subsection{Human-LLM Agreement}
\label{sec:5-2}
\begin{figure}[h]
\centering
\includegraphics[width=0.85\textwidth]{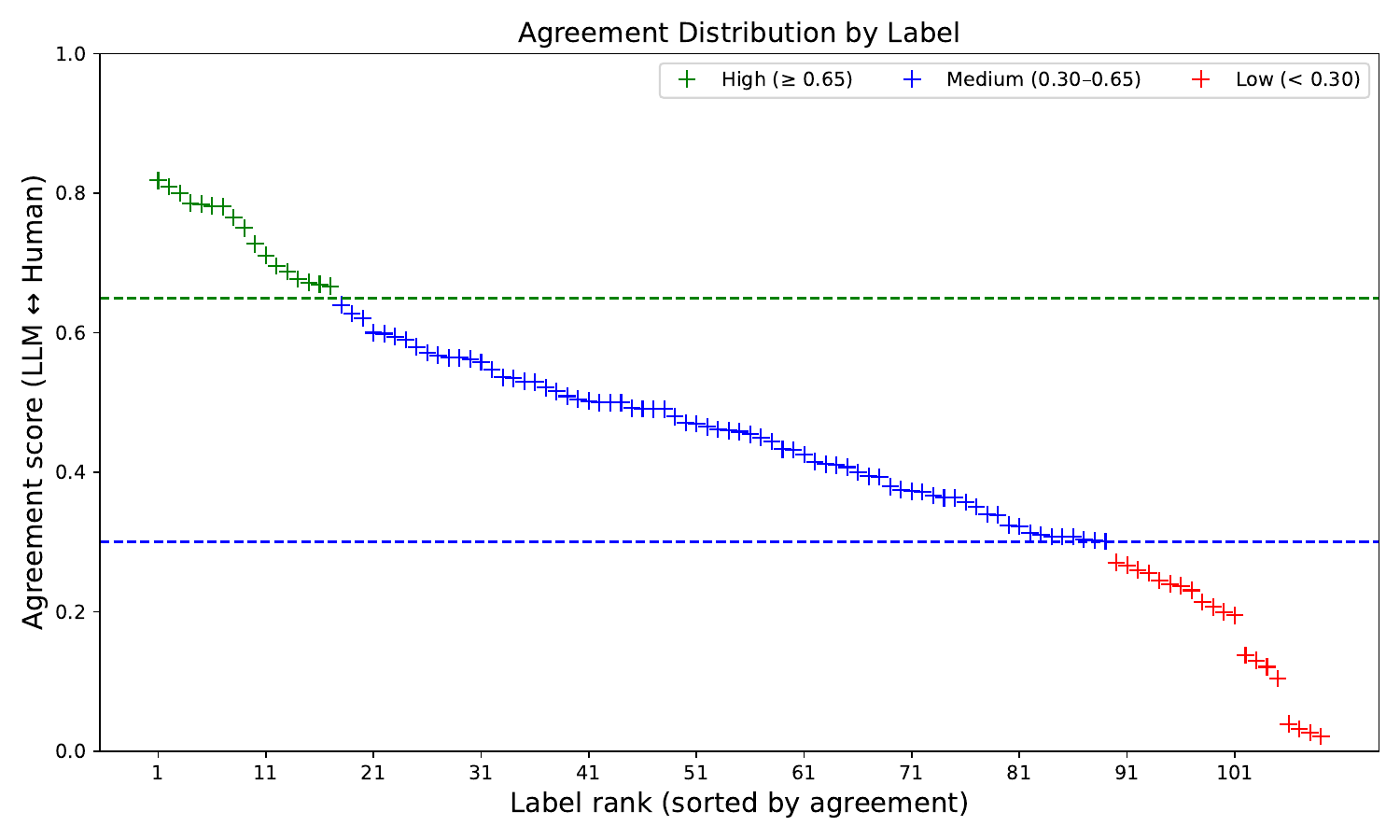}
\caption{\small Human vs LLM agreement score distribution. High agreement scores are shown in green, medium scores in blue, and low scores in red.}
\label{fig:4}
\end{figure}

Figure~\ref{fig:4} presents the distribution of the label-wise agreement scores. The agreement scores shown in the graph are quite dispersed, ranging from high to low, reflecting the overall average level of agreement between human respondents and the LLM. Given that policy indicators span a wide range of dimensions, achieving consistently high agreement is challenging and depends on interpretation, comprehension, and background knowledge.

Our analysis indicates that policy indicators with clear and unambiguous definitions tend to yield higher levels of agreement. The top three labels, one from each category, are clearly interpretable and are as follows: PI015: \textit{``Indirect financial support\textbar Tax or social contributions relief for firms investing in R\&D and innovation''}, TG21: \textit{``Research and education organisations\textbar Public research institutes''}, TH92: \textit{``Net zero transitions\textbar Net zero transitions in energy''}.
Low agreement mainly arises from abstract or broad labels, overlap between categories, and general terminology in policy definitions. The three labels with the lowest scores, one from each category, are as follows: PI010: \textit{``Direct financial support\textbar Procurement programmes for R\&D and innovation''}, TG25: \textit{``Firms by age\textbar Firms of any age''}, TH16: \textit{``Public research system\textbar Public research debates''}.
These patterns highlight that the clarity and specificity of policy indicator definitions play a decisive role in shaping the degree of agreement between human respondents and the LLM.

\subsection{Cross-Validation}
%The OECD survey data consist of multiple types of indicators, including free-text fields and structured policy labels. Manual extraction of these indicators from long, unstructured documents labor-intensive and prone to error. Moreover, 
%The LLM-generated survey responses cannot be fully validated due to the absence of gold standard reference. To address this limitation, we validate LLM-responses against structured policy labels using k-fold cross-validation. For this purpose, we fine-tuned both encoder- and decoder-based models with k set to five. Open-source causal models are fine-tuned using LoRA adapters \cite{hu2022lora}. 
The cross-validation results are reported using micro F1 scores, as shown in Table~\ref{tab:4}. The experiments were conducted with several open- and closed-source models.

\begin{table}[h]
\small
\centering
\caption{\small K-fold cross-validation micro F1 scores for LLM-generated policy indicators.}
\setlength\tabcolsep{10pt}
\begin{tabular}{llccc}
\hline
%&\multicolumn{3}{c}{\textbf{Category-wise (average)}} & \multicolumn{3}{c}{\textbf{Accumulated}} \\
Sr.\# & \textbf{Model}  & \textbf{Precision} & \textbf{Recall} &  \textbf{F1} \\
\hline
1 & RoBERTa-large \cite{liu2019roberta} & 86.70 & 66.61 & 75.33 \\
2 & BigBird-RoBERTa-base \cite{zaheer2020big} & 87.04 & 69.25 & 77.12 \\
3 & BigBird-RoBERTa-large \cite{zaheer2020big} & 88.43 & 76.01 & 81.74 \\
\hline
4 & Llama-3.1-8B-Instruct \cite{grattafiori2024llama3herdmodels} & 82.02 & 90.79 & 86.18 \\
5 & Mistral-7B-Instruct-v0.3 \cite{mistral2023instructv03} &92.91&90.90 &91.89 \\
6 & GPT-OSS-20B \cite{openai2025gptoss120bgptoss20bmodel} & 97.63 & 97.83 & 97.73 \\
\hline
7 & GPT-3.5-Turbo 16k \cite{openai2023gpt35turbo16k} & 98.33 & 97.62 & 97.98 \\
8 & GPT-4o 128k \cite{openai2024gpt4o} & 98.14 & 98.04 & 98.09 \\
\hline
\end{tabular}
\label{tab:4}
\end{table}

%Cross-validation is performed by fine-tuning open-source and closed-source models. 
RoBERTa achieved moderate F-scores, primarily due to its limited input length, whereas its BigBird variants performed better due to their extended input length and block-sparse attention mechanism. In contrast, causal models demonstrated a stronger ability to capture information from longer documents and achieved higher F-scores in multi-label classification. These results highlight the importance of model architecture and context length in determining performance on complex policy classification tasks.

\subsection{Discussion}
\label{sec:5-1}

The findings of this study demonstrate that LLMs can act as effective ``artificial respondents'' for the OECD STIP Compass, offering both efficiency and depth in survey data collection. The comparison between human-generated and LLM-generated responses highlights strong complementarities rather than simple substitution. Specifically, while LLMs tend to provide more detailed procedural and descriptive accounts, human experts emphasize the contextual and societal implications of policy initiatives. 

The overlap analysis shows that LLMs achieve high overlap in structured indicators, with agreement levels of 84--95\% across policy instruments, target groups, and themes. However, in free-text fields such as initiative descriptions and objectives, divergences remain. Only 1.19\% of the cases reached full overlap, while the majority (74.05\%) demonstrated high but not identical overlap. 
The qualitative interpretation of these divergences suggests that they are mainly associated with differences in granularity, framing, and source dependence. LLM-generated responses tend to remain close to explicit source content and provide more procedural detail, whereas human respondents more often compress information and incorporate contextual interpretation. This reinforces the value of a hybrid workflow in which LLMs support initial drafting and pre-filling, while human experts validate objectives and adjust contextual framing.

These patterns highlight the potential of hybrid approaches: LLMs can reduce reporting burdens by pre-filling surveys with structured and detailed information, while human experts can refine, contextualize, and interpret these responses. Nevertheless, risks remain. Over-reliance on synthetic LLM outputs may lead to biases, redundancy, or ``model collapse'' if such outputs are recursively integrated into training data. Ensuring continuous human oversight and triangulation with original sources is thus essential for the long-term integrity of international policy monitoring. In practical deployment, these risks can be mitigated through source-grounded prompting, evidence-based validation, periodic human audits, and avoidance of recursive reuse of unverified AI-generated outputs as training or reference data. These safeguards are especially important for free-text fields, where hallucination and framing bias are more likely than in constrained taxonomy selection.

Overall, the evidence supports the viability of integrating AI into the STIP Compass workflow. Doing so would not only improve scalability and reduce costs but also enhance the descriptive richness of policy monitoring---provided safeguards are in place to preserve contextual accuracy and mitigate systemic biases.

\section{Conclusion}
\label{sec:6}
This study introduces an integrated LLM-based pipeline for policy monitoring within the OECD STIP Compass, demonstrating that AI can complement human expertise in large-scale international surveys. By leveraging long-context in-context learning and a secondary validation layer, the approach achieves high overlap with human-generated responses across structured indicators, while providing additional procedural detail in free-text fields. The results highlight three key findings:
\begin{enumerate}
    \item \textbf{Efficiency gains} -- LLMs can reduce manual reporting burdens by pre-filling structured survey categories.
    \item \textbf{Complementary perspectives} -- LLMs enrich the descriptive layer of policy
    initiatives, while human respondents provide necessary contextualization and societal framing.
    \item \textbf{Scalability with safeguards} -- Hybrid human-AI systems can improve international
    policy intelligence, but careful oversight is required to address risks of bias, redundancy, and
    over-reliance on synthetic outputs.
\end{enumerate}

Future work should expand the scope beyond the six pilot countries, refine validation mechanisms, and explore how human-AI collaboration can be systematically embedded into STI policy monitoring. Ultimately, the integration of LLMs into the STIP Compass marks a step toward more scalable, consistent, and timely global policy intelligence, paving the way for evidence-based innovation governance at the international level.

%\textbf{Use of LLMs:} We acknowledge the use of ChatGPT-5 for writing assistance, grammar polishing, and improving clarity.

\begin{credits}
\subsubsection{\ackname} We acknowledge the use of ChatGPT-5 for language editing, grammar polishing, and improving clarity.

\subsubsection{\discintname}
The authors declare that they have no conflict of interest.
\end{credits}
\bibliographystyle{splncs04}
\bibliography{stip_compass}

\newpage
\appendix

\section{Hyperparameters}
\label{app:a}

\begin{table}[h]
\small
\centering
\caption{Hyperparameters for masked (encoder) models}
\begin{tabular}{ll}
\hline
\textbf{Setting} & \textbf{Value} \\
\hline
%Model & facebook/xlm-roberta-xxl \\
Max sequence length & 512 / 1600 \\
Batch size (train/eval) & 8 / 8 \\
Learning rate & 3e-5 \\
Epochs & 40 \\
Folds & 5 \\
Weight decay & 0.01 \\
Precision & FP16 \\
Eval/save strategy & Per epoch; best model (f1\_micro) \\
Threshold (sigmoid) & 0.5 \\
Label filter & Frequency $\geq$ 5 \\
\hline
\end{tabular}
\end{table}

\begin{table}[h]
\small
\centering
\caption{Hyperparameters for causal models}
\begin{tabular}{ll}
\hline
\textbf{Setting} & \textbf{Value} \\
\hline
%Model & meta-llama/Llama-3.1-8B-Instruct \\
Max input length & 7500 \\
Max new tokens & 200 \\
Batch size (train/eval) & 2 / 2 \\
Gradient accumulation & 8 (effective batch $\approx$ 16) \\
Learning rate & 2e-5 \\
Epochs & 4 \\
Folds & 4 \\
Scheduler & Cosine; warmup ratio 0.03 \\
Precision & bfloat16 \\
Data collator & Causal LM (no MLM) \\
Label filter & Frequency $\geq$ 5 \\
\hline
\end{tabular}
\end{table}

\begin{table}[h]
\small
\centering
\caption{LoRA adapter hyperparameters (causal models)}
\begin{tabular}{ll}
\hline
\textbf{Parameter} & \textbf{Value} \\
\hline
$r$ (rank) & 8 \\
lora\_alpha & 32 \\
lora\_dropout & 0.05 \\
bias & none \\
task\_type & CAUSAL\_LM \\
Target modules & q\_proj, k\_proj, v\_proj, o\_proj \\
\hline
\end{tabular}
\end{table}

\section{Prompts}

\subsection{Validation Prompt}
\label{app:a-1}
\begin{quote}
{\small Evaluate the Response against the given Instructions and Text. Provide a 0/1 assessment for the following dimensions: `evidenced' and `relevant'. 
Use the following criteria to assess `evidenced': Is the Response evidenced in the Text?\\
0 - No, there is no evidence supporting the Response in the Text.\\
1 - Yes, there is evidence supporting the Response in the Text.\\
Use the following criteria to assess `relevant': Is the Response relevant to the Instructions?\\
0 - No, the Response is not relevant to the Instructions (the Response does not follow or answer the Instructions).\\ 
1 - Yes, the Response is relevant to the Instructions (the Response does follow or answer the Instructions).\\
Structure your evaluation in a JSON format with two keys: `evidenced' and `relevant'. `evidenced' should be the 0/1 assessment for whether the response is evidenced in the text, and `relevant' should be the 0/1 assessment for whether the response follows the instructions. Do not elaborate or provide any further explanation.\\
Example JSON structure:\\
\{\\
  ``evidenced'': 1,\\
  ``relevant'': 1\\
\}\\
Instructions: ``+ message +'' Response: + material}
\end{quote}

\subsection{Free-Text Evaluation Prompt}
\label{app:a-2}
\begin{quote}
{\small "You are given two sets of policy initiative documents: one is [human assessment] and the other is [LLM assessment]. Your task is to analyze these documents to understand their similarities, overlaps, and differences using quantifiable metrics. Determine the level of overlap and choose only one category from "Full overlap", "High overlap", "Low overlap", or "No overlap". Provide the response in the following JSON format with an appropriate category and observation.\\
Example 1:\\
\{\\
      "Full overlap": \{\\
        "Observation": "Both assessments agreed entirely on the need for increased funding for education."\}\\
    \}\\
Example 2:\\
    \{\\
      "High overlap": \{\\
        "Observation": "Both assessments focus on promoting RDI activities, increasing competitiveness, and attracting foreign investments. However, the LLM assessment provides a more detailed breakdown of these objectives."\}\\
    \}\\
Example 3:\\
    \{\\
      "Low overlap": \{\\
        "Observation": "The LLM emphasized renewable energy incentives more than the human assessment."\}\\
    \}\\
    Example 4:\\
    \{\\
      "No overlap": \{\\
        "Observation": "The human assessment discussed healthcare reforms, which were not mentioned by the LLM."\}\\
    \}\\
}
\end{quote}

\subsection{Pre-Filling Prompts}
\label{app:a-3}

\begin{quote}
\textit{Identification}\\
{\small Using the information provided, determine whether [REPLACE] is discussed. Respond 1 if yes and 0 if no. If you cannot determine whether the text contains information about [REPLACE], respond 99 and do not elaborate. Only respond with 0, 1, or 99.
}
\end{quote}

\begin{quote}
\textit{Description}\\
{\small Act as an expert policy analyst. Based on the given policy-related text material provide a short description in English of [REPLACE] in sentence format, not exceeding 100 words. Avoid using jargon; be concise and clear, delivering only information retrieved from the text. If there is no discussion or mention of the topic, respond with "No information" and do not elaborate. Provide the response in a JSON format where the suggested name or theme is the key and the description is the value.\\
Example JSON structure:\\
\{\\
  "description": "Description of the relevant policy content in a clear and concise sentence."\\
\}
}
\end{quote}

\begin{quote}
\textit{Objectives}\\
{\small Act as a policy expert identify the [REPLACE] initiative's objectives discussed. Structure your response in JSON format with two keys: 'objective' and 'description'. 'objective' should be a short title of the objective and 'description' should be a brief explanation of the objective, not exceeding 100 words. Both should be provided in English. If you did not find any objective just indicate "n.a."\\
Example JSON structure:\\
\{\\
  "objective": "Enhance International Profile",\\
  "description": "To give public research institutes a higher profile in the international context by providing funding for international collaboration."\\
\}
}
\end{quote}

\begin{quote}
\textit{Start date}\\
{\small Using the information provided, which is a policy-related document, your task is to determine the starting date of the [REPLACE] initiative if mentioned. Structure your answers as a JSON file where the key is the date which can be year and month and the value is the description of what the date refers to in English. Avoid using jargon; be concise and clear, delivering only information retrieved from the text. If there is no discussion or mention of the topic, respond with "n.a." and do not elaborate.\\
Example JSON structure:\\
\{\\
    "2024-12": "Start date of the new environmental regulation initiative.",\\
  "2023-06": "Launch date of the public health awareness campaign.",\\
  "n.a.": "No starting date mentioned for the initiative."\\
\}
}
\end{quote}

\begin{quote}
\textit{Policy instruments}\\
{\small Consider yourself an expert policy analyst tasked with reading documents related to Science Technology and Innovation (STI) policy. Your goal is to comprehend and identify which of the below instrument(s) the [REPLACE] initiative is relevant to and then assign them to relevant policy instrument types based on the information within the given policy instrument labels and policy instrument definitions. These categories are provided to you in a JSON file with keys such as "policyInstrumentID" for the ID of that policy instrument, "label" for the descriptive label of the policy instrument, and "definition" for the description of what the policy instrument is and what it relates to. Based on the text given to you, identify which definitions of the given policy instrument types fit and are mentioned in the text, and identify the Policy Instrument ID with your short justification in English. A text can be relevant to one or more Policy Instrument ID, return only relevant matches. You should structure your response as a JSON array where each object contains "PolicyInstrumentID" as the key for the policy instrument ID and "reason" as the key for your reasoning. If you don’t find any relevant Policy Instrument, just say "n.a."\\
Example response format:\\
\begin{verbatim}
[
  {
    "PolicyInstrumentID": "PI019",
    "reason": "reasoning..."
  },
  {
    "PolicyInstrumentID": "PI020",
    "reason": "reasoning..."
  }
]
\end{verbatim}
If no relevant Policy instrument is found, the response should be:\\
\{\\
 "n.a.": "No relevant Policy instrument found."\\
\}\\
\\
Here are the Policy Instrument Types, labels, and their definitions:\\
 PI\_Code: PI024, Label: Governance|Strategies, agendas and plans,\\
 PI\_Code: PI030, Label: Governance|Creation or reform of governance structure or public body,\\
 PI\_Code: PI031, Label: Governance|Policy intelligence (e.g. evaluations, benchmarking and forecasts),\\
 PI\_Code: PI025, Label: Governance|Formal consultation of stakeholders or experts,\\
 PI\_Code: PI026, Label: Governance|Horizontal STI coordination bodies,\\
 PI\_Code: PI033, Label: Governance|Regulatory oversight and ethical advice bodies,\\
 PI\_Code: PI027, Label: Governance|Standards and certification for technology development and adoption,\\
 PI\_Code: PI028, Label: Governance|Public awareness campaigns and other outreach activities,\\
 PI\_Code: PI006, Label: Direct financial support|Institutional funding for public research,\\
 PI\_Code: PI007, Label: Direct financial support|Project grants for public research,\\
 PI\_Code: PI008, Label: Direct financial support|Grants for business R\&D and innovation,\\
 PI\_Code: PI009, Label: Direct financial support|Centres of excellence grants,\\
 PI\_Code: PI010, Label: Direct financial support|Procurement programmes for R\&D and innovation,\\
 PI\_Code: PI011, Label: Direct financial support|Fellowships and postgraduate loans and scholarships,\\
 PI\_Code: PI012, Label: Direct financial support|Loans and credits for innovation in firms,\\
 PI\_Code: PI013, Label: Direct financial support|Equity financing,\\
 PI\_Code: PI014, Label: Direct financial support|Innovation vouchers,\\
 PI\_Code: PI015, Label: Indirect financial support|Tax or social contributions relief for firms investing in R\&D and innovation,\\
 PI\_Code: PI016, Label: Indirect financial support|Tax relief for individuals supporting R\&D and innovation,\\
 PI\_Code: PI029, Label: Indirect financial support|Debt guarantees and risk sharing schemes,\\
 PI\_Code: PI021, Label: Collaborative infrastructures (soft and physical)|Networking and collaborative platforms,\\
PI\_Code: PI022, Label: Collaborative infrastructures (soft and physical)|Dedicated support to research and technical infrastructures,\\
 PI\_Code: PI023, Label: Collaborative infrastructures (soft and physical)|Information services and access to datasets,\\
PI\_Code: PI017, Label: Guidance, regulation and incentives|Technology extension and business advisory services,\\
PI\_Code: PI032, Label: Guidance, regulation and incentives|Science and technology regulation and soft law,\\
 PI\_Code: PI018, Label: Guidance, regulation and incentives|Labour mobility regulation and incentives\\
 PI\_Code: PI019, Label: Guidance, regulation and incentives|Intellectual property regulation and incentives,\\
PI\_Code: PI020, Label: Guidance, regulation and incentives|Science and innovation challenges, prizes and awards,\\

}   %end tt
\end{quote}

\begin{quote}
\textit{Policy target groups}\\
{\small Consider yourself an expert policy analyst tasked with reading documents related to Science Technology and Innovation (STI) policy. Your goal is to comprehend and identify which of the below target group(s) the [REPLACE] initiative is relevant to and then assign them to relevant target groups based on the information within the given target group labels. These categories are provided to you in a JSON file with keys such as "target group code" for the ID of that target group, and "target group name" for the descriptive label of the target group. Based on the text given to you, identify which of the given target groups types fit and are mentioned in the text, and identify the target group code with your short justification in English. A text can be relevant to one or more target group, return only relevant matches. You should structure your response as a JSON array where each object contains "TargetGroupID" as the key for the target group ID and "reason" as the key for your reasoning. If you don't find any relevant target group, just say "n.a."\\
Example response format:
\begin{verbatim}
[
  {
    "TargetGroupID": "TG20",
    "reason": "reasoning..."
  },
  {
    "TargetGroupID": "TG9",
    "reason": "reasoning..."
  }
]
\end{verbatim}
If no relevant target group is found, the response should be:\\
TG\_Code : TG20, Label : Research and education organisations|Higher education institutes \\
TG\_Code : TG21, Label : Research and education organisations|Public research institutes \\
TG\_Code : TG22, Label : Research and education organisations|Private research and development lab 
TG\_Code : TG9, Label : Researchers, students and teachers|Established researchers \\
TG\_Code : TG11, Label : Researchers, students and teachers|Postdocs and other early-career researchers \\
TG\_Code : TG41, Label : Researchers, students and teachers|Programme managers and other research support staff \\
TG\_Code : TG10, Label : Researchers, students and teachers|Undergraduate and master students \\
TG\_Code : TG38, Label : Researchers, students and teachers|Secondary education students \\
TG\_Code : TG12, Label : Researchers, students and teachers|PhD students \\
TG\_Code : TG13, Label : Researchers, students and teachers|Teachers \\
TG\_Code : TG29, Label : Firms by size|Firms of any size \\
TG\_Code : TG30, Label : Firms by size|Micro-enterprises \\
TG\_Code : TG31, Label : Firms by size|SMEs \\
TG\_Code : TG32, Label : Firms by size|Large firms \\
TG\_Code : TG33, Label : Firms by size|Multinational enterprises \\
TG\_Code : TG25, Label : Firms by age|Firms of any age \\
TG\_Code : TG26, Label : Firms by age|Nascent firms (0 to less than 1 year old) \\
TG\_Code : TG27, Label : Firms by age|Young firms (1 to 5 years old) \\
TG\_Code : TG28, Label : Firms by age|Established firms (more than 5 years old) \\
TG\_Code : TG34, Label : Intermediaries|Incubators, accelerators, science parks or technoparks 
TG\_Code : TG35, Label : Intermediaries|Technology transfer offices \\
TG\_Code : TG36, Label : Intermediaries|Industry associations \\
TG\_Code : TG37, Label : Intermediaries|Academic societies / academies \\
TG\_Code : TG42, Label : Intermediaries|Non-governmental organisations (NGOs) \\
TG\_Code : TG40, Label : Governmental entities|International entity \\
TG\_Code : TG23, Label : Governmental entities|National government \\
TG\_Code : TG24, Label : Governmental entities|Subnational government \\
TG\_Code : TG18, Label : Economic actors (individuals)|Entrepreneurs \\
TG\_Code : TG17, Label : Economic actors (individuals)|Private investors \\
TG\_Code : TG19, Label : Economic actors (individuals)|Labour force in general \\
TG\_Code : TG14, Label : Social groups especially emphasised|Women \\
TG\_Code : TG15, Label : Social groups especially emphasised|Disadvantaged and excluded groups 
TG\_Code : TG16, Label : Social groups especially emphasised|Civil society \\

}   %end tt
\end{quote}

\begin{quote}
\textit{Policy themes}\\
{\small Consider yourself an expert policy analyst tasked with reading documents related to Science, Technology, and Innovation (STI) policy. Your goal is to comprehend and identify which of the below policy themes the [REPLACE] initiative is relevant to and assign the appropriate policy themes based on the information within the given policy theme labels and policy theme relevancy guiding questions. These categories are provided to you in a JSON file where you can find the guiding "question" to ask before assigning the policy theme. Each policy theme includes a "label" and a "code”. Based on these guidelines, identify which policy theme "code” fits the given policy theme name and related question, and provide a short justification for your selection in English. A text can be relevant to one or more policy theme, return only relevant matches. Structure your response as a JSON array where each object contains "PolicyThemeCode" as the key for the policy theme and "reason" as the key for your reasoning. If you don’t find any relevant policy theme, just say "n.a."\\
Example response format:
\begin{verbatim}
[
  {
    "PolicyThemeCode": "TH26",
    "reason": "reasoning..."
  },
  {
    "PolicyThemeCode": "TH30",
    "reason": "reasoning..."
  }
]
\end{verbatim}
If no relevant policy theme is found, the response should be:\\
\{\\
  "n.a.": "No relevant policy theme found."\\
\}\\
Here are the policy theme codes, labels, and their defining questions in a structured JSON file:\\
TH\_Code : TH11, Label: Governance|Governance debates,\\
TH\_Code : TH13, Label: Governance|STI plan or strategy,\\
TH\_Code : TH9, Label: Governance|Horizontal policy coordination,\\
TH\_Code : TH14, Label: Governance|Strategic policy intelligence,\\
TH\_Code : TH15, Label: Governance|Evaluation and impact assessment,\\
TH\_Code : TH63, Label: Governance|International STI governance policy,\\
TH\_Code : TH16, Label: Public research system|Public research debates,\\
TH\_Code : TH18, Label: Public research system|Public research strategies,\\
TH\_Code : TH19, Label: Public research system|Competitive research funding,\\
TH\_Code : TH20, Label: Public research system|Non-competitive research funding,\\
TH\_Code : TH27, Label: Public research system|Third-party funding,\\
TH\_Code : TH22, Label: Public research system|Structural change in the public research system,\\
TH\_Code : TH106, Label: Public research system|Digital transformation of research-performing organisations,\\
TH\_Code : TH107, Label: Public research system|Open and enhanced access to publications,\\
TH\_Code : TH108, Label: Public research system|Open and enhanced access to research data,\\
TH\_Code : TH24, Label: Public research system|Research and technology infrastructures,\\
TH\_Code : TH25, Label: Public research system|Internationalisation in public research,\\
TH\_Code : TH26, Label: Public research system|Cross-disciplinary research,\\
TH\_Code : TH23, Label: Public research system|High-risk high-reward research,\\
TH\_Code : TH21, Label: Public research system|Research integrity and reproducibility,\\
TH\_Code : TH109, Label: Public research system|Research security,\\
TH\_Code : TH28, Label: Innovation in firms and innovative entrepreneurship|Business innovation policy debates,\\
TH\_Code : TH30, Label: Innovation in firms and innovative entrepreneurship|Business innovation policy strategies,\\
TH\_Code : TH31, Label: Innovation in firms and innovative entrepreneurship|Financial support to business R\&D and innovation,\\
TH\_Code : TH32, Label: Innovation in firms and innovative entrepreneurship|Non-financial support to business R\&D and innovation,\\
TH\_Code : TH38, Label: Innovation in firms and innovative entrepreneurship|Access to finance for innovation,\\
TH\_Code : TH34, Label: Innovation in firms and innovative entrepreneurship|Entrepreneurship capabilities and culture,\\
TH\_Code : TH33, Label: Innovation in firms and innovative entrepreneurship|Stimulating demand for innovation and market creation,\\
TH\_Code : TH82, Label: Innovation in firms and innovative entrepreneurship|Digital transformation of firms,\\
TH\_Code : TH36, Label: Innovation in firms and innovative entrepreneurship|Foreign direct investment,\\
TH\_Code : TH35, Label: Innovation in firms and innovative entrepreneurship|Targeted support to SMEs and young innovative enterprises,\\
TH\_Code : TH39, Label: Knowledge exchange and co-creation|Knowledge exchange and co-creation debates,\\
TH\_Code : TH41, Label: Knowledge exchange and co-creation|Knowledge exchange and co-creation strategies,\\
TH\_Code : TH42, Label: Knowledge exchange and co-creation|Collaborative research and innovation,\\
TH\_Code : TH47, Label: Knowledge exchange and co-creation|Cluster policies,\\
TH\_Code : TH43, Label: Knowledge exchange and co-creation|Commercialisation of public research results,\\
TH\_Code : TH44, Label: Knowledge exchange and co-creation|Inter-sectoral mobility,\\
TH\_Code : TH46, Label: Knowledge exchange and co-creation|Intellectual property rights in public research,\\
TH\_Code : TH48, Label: Human resources for research and innovation|STI human resources debates,\\
TH\_Code : TH50, Label: Human resources for research and innovation|STI human resources strategies,\\
TH\_Code : TH51, Label: Human resources for research and innovation|STEM skills,\\
TH\_Code : TH52, Label: Human resources for research and innovation|Doctoral and postdoctoral researchers,\\
TH\_Code : TH53, Label: Human resources for research and innovation|Research careers,\\
TH\_Code : TH55, Label: Human resources for research and innovation|International mobility of human resources,\\
TH\_Code : TH54, Label: Human resources for research and innovation|Equity, diversity and inclusion (EDI),\\
TH\_Code : TH56, Label: Research and innovation for society|Policy debates on innovation for societal challenges,\\
TH\_Code : TH58, Label: Research and innovation for society|Research and innovation for society strategy,\\
TH\_Code : TH91, Label: Research and innovation for society|Mission-oriented innovation policies,\\
TH\_Code : TH89, Label: Research and innovation for society|Ethics of emerging technologies,\\
TH\_Code : TH61, Label: Research and innovation for society|Research and innovation for developing countries,\\
TH\_Code : TH65, Label: Research and innovation for society|Multi-stakeholder engagement,\\
TH\_Code : TH66, Label: Research and innovation for society|Science, technology and innovation culture,\\
TH\_Code : TH101, Label: Net zero transitions|Net zero transitions policy debates,\\
TH\_Code : TH102, Label: Net zero transitions|Government capabilities for net zero transitions,\\
TH\_Code : TH92, Label: Net zero transitions|Net zero transitions in energy,\\
TH\_Code : TH103, Label: Net zero transitions|Net zero transitions in transport and mobility,\\
TH\_Code : TH104, Label: Net zero transitions|Net zero transitions in food and agriculture,\\
TH\_Code : TH105, Label: Net zero transitions|STI policies for net zero,

}   %end tt
\end{quote}

\begin{quote}
\textit{Budget}\\
{\small Using the information provided, your task is to determine if there is any monetary information such as budget or expenditure related to the [REPLACE] initiative. Make your response in English. If there is no information, respond with "No information" and do not elaborate. Answer in English and provide your answers in a JSON structured format.\\
Example JSON response:\\
```json[\\
\{ \\
"monetaryInformation": "Budget of \$10 million allocated for research and development." \\
\},\\
\{\\ 
"monetaryInformation": "Budget of \$5 million allocated for implementation."\\ 
\}\\
]\\
```\\
If no monetary information is found, the response should be:\\
```json\\
\{\\
  "monetaryInformation": "No information"\\
\}\\
```
}   %end tt
\end{quote}

\begin{quote}
\textit{Evaluation report}\\
{\small Using the information provided, your task is to determine if the [REPLACE] initiative has been evaluated and if an evaluation report exists. If evaluation is not mentioned, respond with "No information" and do not elaborate. Structure your information as a JSON file where the key is "evaluation name” and the value is "the information of the found evaluation” in English. Avoid using jargon; be concise and clear, delivering only information retrieved from the text.\\
Example JSON response:\\
```json [\\
\{\\
  "evaluationName": "Mid-Term Evaluation Report",\\
  "information": "The mid-term evaluation report conducted in 2023 assesses the effectiveness and impact of the policy initiative."\\
\}\\
]\\
```\\
If no evaluation information is found, the response should be:\\
```json\\
\{\\
  "evaluationName": "n.a.",\\
  "information": "No information"\\
\}\\
```\\
}   %end tt
\end{quote}

\subsection{System Prompt for Fine-tuning Causal Models}
\label{app:a-4}
\begin{quote}
{\small You are an AI assistant trained to classify text about Science, Technology, and Innovation (STI) policy. Your task is to identify the most relevant category labels from the three dimensions below:\\
1. Policy Instruments (PI)\\
2. Policy Target Groups (TG)\\
3. Policy Themes (TH)\\
Each category has a list of definitions provided. Based on the input text, identify and return only the **Labels** of items that are clearly relevant to the content. Your answer should be only a **flat list of matching labels**. DO NOT provide any additional text.\\
Policy Instruments (PI) labels and definitions:\\
\{
pi\_definitions
\}\\
Policy Target Groups (TG) labels and definitions:\\
\{
tg\_definitions
\}\\
Policy Themes (TH) labels and definitions:\\
\{
th\_definitions
\}\\
}
\end{quote}

\end{document}